\documentclass[conference]{IEEEtran}

\usepackage{cite}
\usepackage{amsmath,amssymb,amsfonts}
\usepackage{mathtools}
\usepackage{algorithmic}
\usepackage{graphicx}
\usepackage{comment}
\usepackage{textcomp}
\usepackage{xcolor}
\usepackage{xfrac}
\usepackage{units}
\usepackage{threeparttable}
\usepackage{booktabs}
\usepackage{subfig}
\usepackage{wrapfig}
\usepackage{hyperref}

\makeatletter \newcommand{\linebreakand}{\end{@IEEEauthorhalign}
  \hfill\mbox{}\par
  \mbox{}\hfill\begin{@IEEEauthorhalign}
}
\makeatother \newcommand{\NR}{Newton-Raphson}

\definecolor{rwth-blue}{cmyk}{1,.5,0,0}\colorlet{rwth-lblue}{rwth-blue!50}\colorlet{rwth-llblue}{rwth-blue!25}
\definecolor{rwth-violet}{cmyk}{.6,.6,0,0}\colorlet{rwth-lviolet}{rwth-violet!50}\colorlet{rwth-llviolet}{rwth-violet!25}
\definecolor{rwth-purple}{cmyk}{.7,1,.35,.15}\colorlet{rwth-lpurple}{rwth-purple!50}\colorlet{rwth-llpurple}{rwth-purple!25}
\definecolor{rwth-carmine}{cmyk}{.25,1,.7,.2}\colorlet{rwth-lcarmine}{rwth-carmine!50}\colorlet{rwth-llcarmine}{rwth-carmine!25}
\definecolor{rwth-red}{cmyk}{.15,1,1,0}\colorlet{rwth-lred}{rwth-red!50}\colorlet{rwth-llred}{rwth-red!25}
\definecolor{rwth-magenta}{cmyk}{0,1,.25,0}\colorlet{rwth-lmagenta}{rwth-magenta!50}\colorlet{rwth-llmagenta}{rwth-magenta!25}
\definecolor{rwth-orange}{cmyk}{0,.4,1,0}\colorlet{rwth-lorange}{rwth-orange!50}\colorlet{rwth-llorange}{rwth-orange!25}
\definecolor{rwth-yellow}{cmyk}{0,0,1,0}\colorlet{rwth-lyellow}{rwth-yellow!50}\colorlet{rwth-llyellow}{rwth-yellow!25}
\definecolor{rwth-grass}{cmyk}{.35,0,1,0}\colorlet{rwth-lgrass}{rwth-grass!50}\colorlet{rwth-llgrass}{rwth-grass!25}
\definecolor{rwth-green}{cmyk}{.7,0,1,0}\colorlet{rwth-lgreen}{rwth-green!50}\colorlet{rwth-llgreen}{rwth-green!25}
\definecolor{rwth-cyan}{cmyk}{1,0,.4,0}\colorlet{rwth-lcyan}{rwth-cyan!50}\colorlet{rwth-llcyan}{rwth-cyan!25}
\definecolor{rwth-teal}{cmyk}{1,.3,.5,.3}\colorlet{rwth-lteal}{rwth-teal!50}\colorlet{rwth-llteal}{rwth-teal!25}
\definecolor{rwth-gold}{cmyk}{.35,.46,.7,.35}
\definecolor{rwth-silver}{cmyk}{.39,.31,.32,.14}

\def\BibTeX{{\rm B\kern-.05em{\sc i\kern-.025em b}\kern-.08em
    T\kern-.1667em\lower.7ex\hbox{E}\kern-.125emX}}
    
\begin{document}

\title{Solving AC Power Flow with Graph Neural Networks under Realistic Constraints}

\author{\IEEEauthorblockN{Luis Böttcher$^*$}
\IEEEauthorblockA{\textit{IAEW} \\
\textit{RWTH Aachen University}\\
Aachen, Germany \\
l.boettcher@iaew.rwth-aachen.de}
\and
\IEEEauthorblockN{Hinrikus Wolf$^*$}
\IEEEauthorblockA{\textit{Computer Science} \\
\textit{RWTH Aachen University}\\
Aachen, Germany \\
hinrikus@cs.rwth-aachen.de}
\and
\IEEEauthorblockN{Bastian Jung$^*$}
\IEEEauthorblockA{\textit{IAEW} \\
\textit{RWTH Aachen University}\\
Aachen, Germany \\
bastian.jung@rwth-aachen.de}
\linebreakand

\IEEEauthorblockN{Philipp Lutat}
\IEEEauthorblockA{\textit{IAEW} \\
\textit{RWTH Aachen University}\\
Aachen, Germany \\
p.lutat@iaew.rwth-aachen.de}
\and
\IEEEauthorblockN{Marc Trageser}
\IEEEauthorblockA{\textit{IAEW} \\
\textit{RWTH Aachen University}\\
Aachen, Germany \\
m.trageser@iaew.rwth-aachen.de}
\and
\IEEEauthorblockN{Oliver Pohl}
\IEEEauthorblockA{\phantom{\textit{IAEW}} \\
\textit{Schleswig-Holstein Netz AG}\\
Quickborn, Germany \\
oliver.pohl@sh-netz.de}
\linebreakand

\IEEEauthorblockN{Xiaohu Tao}
\IEEEauthorblockA{\phantom{\textit{IAEW}} \\
\textit{E.ON SE}\\
Essen, Germany \\
xiaohu.tao@eon.com}
\and
\IEEEauthorblockN{Andreas Ulbig}
\IEEEauthorblockA{\textit{IAEW} \\
\textit{RWTH Aachen University}\\
Aachen, Germany \\
a.ulbig@iaew.rwth-aachen.de}
\and
\IEEEauthorblockN{Martin Grohe}
\IEEEauthorblockA{\textit{Computer Science} \\
\textit{RWTH Aachen University}\\
Aachen, Germany \\
grohe@cs.rwth-aachen.de}
}
\IEEEoverridecommandlockouts
\IEEEpubid{
\begin{minipage}{\columnwidth}
\centering
DOI: \href{https://doi.org/10.1109/PowerTech55446.2023.10202246}{10.1109/PowerTech55446.2023.10202246} \\
    978-1-6653-8778-8/23/\$31.00~\copyright2023 IEEE 
\end{minipage}
\hspace{\columnsep}\makebox[\columnwidth]{ }
}

\maketitle
{
\def\thefootnote{*}\footnotetext{These authors contributed equally to this work}
}
\IEEEpubidadjcol
\begin{abstract}
In this paper, we propose a graph neural network architecture to solve the AC power flow problem under realistic constraints. 
To ensure a safe and resilient operation of distribution grids, AC power flow calculations are the means of choice to determine grid operating limits or analyze grid asset utilization in planning procedures. In our approach, we demonstrate the development of a framework that uses graph neural networks to learn the physical constraints of the power flow. We present our model architecture on which we perform unsupervised training to learn a general solution of the AC power flow formulation independent of the specific topologies and supply tasks used for training.
Finally, we demonstrate, validate and discuss our results on medium voltage benchmark grids.
In our approach, we focus on the physical and topological properties of distribution grids to provide scalable solutions for real grid topologies. Therefore, we take a data-driven approach, using large and diverse data sets consisting of realistic grid topologies, for the unsupervised training of the AC power flow graph neural network architecture and compare the results to a prior neural architecture and the Newton-Raphson method. Our approach shows a high increase in computation time and good accuracy compared to state-of-the-art solvers. 
It also out-performs that neural solver for power flow in terms of accuracy.
\end{abstract}


\section{Introduction}
The AC power flow determines the steady-state operating condition on a given grid topology with a corresponding supply task, i.e. the electricity demand and production profiles at the different grid nodes. It is the prerequisite and core of grid planning and operation processes. With a growing share of renewable energy sources and new electric loads to be handled by low and medium-voltage grids, the operation and planning of distribution grids will become more challenging.
Consequently, it can be assumed that the passively designed grid infrastructure will be operated actively in the future. In contrast, continuous simulations of the operating states of grids with a high number of nodes and assets will be carried out to ensure an efficient operation of the grid. Accurate solution methods already exist for solving the AC power flow problem (e.g. the \NR{} method). However, computation times scale significantly with increasing grid size and observation horizon. 
For the analysis of transmission grids, simplifications, like the DC power flow and fast decoupled power flow, are sufficiently precise to generate power flow solutions for determining grid congestions. 
However, this is not the case in the underlying distribution grid due to the $R$ to $X$ ratio. Therefore, for the accurate analysis of distribution grids on medium and low voltage levels, the AC power flow has to be determined. 

Solving the AC power flow with graph neural networks (GNN) allows high parallel computing power and can provide shorter computation times for large instances. 
Therefore, this paper presents a physics-aware GNN architecture for AC Power Flow on distribution grids.
It can learn the Power Flow without any ground truth labels, i.e., any prior \NR{} solutions.
The GNN utilises a loss function based on the Power Flow Equations, thus learning the physical constraints of power grids.
By this, the GNN can determine the quality of solutions while training and in inference.
We trained on large and diverse data sets consisting of many artificial and real-world grid topologies.
Further, the model can generalise well to predict a power flow on unseen topologies because of the utilisation of graph neural networks.

The structure of the paper is as follows. Section II  presents the state of the art and points out the contribution of this paper. Section III represents the Background on Power Flow and Graph Neural Networks. In Section IV, the methodology is presented. Section V describes the experiments conducted. The experiments are discussed in Section VI. Finally, the conclusions of the proposed work are given in Section VII.

\section{State of the Art and Contribution}
The viability of using graph neural networks to solve power flow calculations has recently been demonstrated in \cite{Donon.2019} and \cite{Donon.2020}. These publications focus on solving power flows on a transmission grid level. For distribution grids, the focus shifts to the physical characteristics of the distribution system. A review of other applications of GNN in power systems is presented in \cite{Liao.2022}. In addition to power flow calculations, an approach to compute optimal power flows based on graph neural networks is proposed in \cite{Owerko.2019}. This publication also considers solving the power flow constraints on meshed topologies. In \cite{Jeddi.2021} a graph attention mechanism is used to find suitable solutions for the power flow calculation. Another approach is presented in \cite{Hansen.2021} in which the authors develop specialised GNN layers to determine power flows. In \cite{Hu.2021} a data-driven neural approach for computing power flow is presented.

All these approaches are developed, trained and tested on benchmark network models. 
Compared to other approaches, our contribution with this paper lies in the realistic modelling of input training data based on synthetic distribution grid models to adequately model the power flow on distribution grids. Furthermore, we provide an unsupervised machine learning approach using graph neural networks to learn the physical constraints of power grids and solve the AC power flow.

 \section{Background}
This section provides an overview of state of the art on power flow and Graph Neural Network basics.

 \subsection{Power Flow Formulation}
Network congestions, such as the violation of voltage limits and overloading of network resources, are identified by power flow calculations for different feed-in and load situations. In grid operation, the grid-supporting use of flexibility, such as regulating generation plants or controlling consumption devices, can be determined with optimal power flow formulations. Power flow calculations are necessary for grid planning to identify and check possible grid expansion measures to eliminate grid congestion.
The AC power flow problem is defined on a set of nodes and edges together with their corresponding power flow equations:

\begin{equation*}
     P_{i} = \sum_{k=1}^N |V_i||V_k|(G_{ik}\cos\vartheta_{ik}+B_{ik}\sin\vartheta_{ik})
\end{equation*}

\begin{equation*}
    Q_{i} = \sum_{k=1}^N |V_i||V_k|(G_{ik}\sin\vartheta_{ik}-B_{ik}\cos\vartheta_{ik}).
\end{equation*}

Each node is defined by its four properties: real power $P_i$, reactive power $Q_i$, voltage magnitude $V_i$ and $V_k$, and voltage angle $\vartheta_{ik}$. The power lines are characterised by their admittance, separated into their real part $G_{ik}$ and imaginary part $B_{ik}$. The power flow describes the state of a power system with a set of given nodes and edges sufficiently, as long as at least two of the four node variables are known. Due to their operational behaviour, node components (loads or generators) are split into subcategories of different value assignments. Table \ref{tab:bus_types} provides an overview of the node types necessary for a power flow calculation. Detailed Information on the power flow formulation can be found in varying literature such as \cite{Chow.2020}.

 \begin{table}[h]
     \centering
         \caption{Power Flow Bus Types}
     \begin{tabular}{lcccc}
     \toprule
         \textbf{Bus type} & \textbf{$P$}& \textbf{$Q$}& \textbf{$V$} & \textbf{$\vartheta$}  \\
          \midrule
          PQ (load) & Specified & Specified & Unknown & Unknown  \\
          PV (generator) & Specified & Unknown & Specified & Unknown \\
          Slack (balance) & Unknown & Unknown & Specified & Specified \\
          \bottomrule
     \end{tabular}
 
     \label{tab:bus_types}
 \end{table}

 \subsection{Graph Neural Networks}
 In the last few years, GNNs (Graph Neural Networks) have emerged as a standard method for machine learning tasks on graphs \cite{DBLP:conf/iclr/KipfW17, DBLP:conf/aaai/0001RFHLRG19,DBLP:conf/iclr/XuHLJ19, DBLP:journals/tnn/WuPCLZY21}.
They can be used for the analysis of chemical molecules as well as for larger graph structures such as social networks.
 Graph Neural Networks utilise the concept of message passing \cite{DBLP:conf/icml/GilmerSRVD17} as a layer in a neural network.
 In such a layer, messages of latent features are exchanged between the graph nodes, which are then processed by other neural network components.
Message passing can be seen as a communication protocol between nodes in the graph, which uses this protocol repeatedly as part of a single GNN forward pass.
 Gilmer et al. \cite{DBLP:conf/icml/GilmerSRVD17} define message passing as follows.
Let \( \mathbf{G}=(\mathbf{V}, \mathbf{E})\) be a graph.  For every node \(\mathbf{v}_i \in \mathbf{V}\) a message passing layer has a \emph{hidden state}  \(h^t_{\mathbf{G}_i}\) for time stamp \(t\).
 The hidden states are initialised randomly for most applications at time stamp \(t=0\).
Message passing consists of two phases: an update phase and a read-out phase.
In the update phase \((t \rightarrow t+1) \) every node \(\mathbf{v}\) sends its hidden state \(h_{\mathbf{v}}^t\) to its neighbours.
Based on the received messages a node \(\mathbf{w}\) updates its hidden states by
\begin{equation*}
 m_{\mathbf{v}}^{t+1} = \sum_{\mathbf{w}\in N(v)} M^t(h_{\mathbf{v}}^t, h_{\mathbf{w}}^t, e_{\mathbf{v}\mathbf{w}})
\end{equation*}

and 
\begin{equation*}
 h_{\mathbf{v}}^{t+1} = U^t(h_{\mathbf{v}}^t, m_{\mathbf{v}}^{t})
\end{equation*}

 with \(e_{\mathbf{v}\mathbf{w}}\) the edge label of edge \((\mathbf{v},\mathbf{w})\) and \(M, U\) trainable functions, i.e., linear layers with an activation function.
The read-out phase can be seen as an interface to the successive layers of the neural networks.
The read-out function \( \hat{y} = R(h_{\mathbf{v}}^t \mid\mathbf{v} \in \mathbf{V})\) is executed, with \(R\) being a trainable function.
Various extensions have been made to this definition, e.g., by adding LSTM cells.
In this paper, we train unsupervised GNNs in an approach inspired by \cite{DBLP:journals/frai/TonshoffRWG20}.

 \section{Methodology}
In general, we can separate our approach into the data preparation process, the loss function used for training and the model architecture.
\subsection{Grid Data Generation} 
\begin{figure}[h]
    \centering
    \subfloat[\centering Low-voltage topology]{{\includegraphics[width=.4\linewidth]{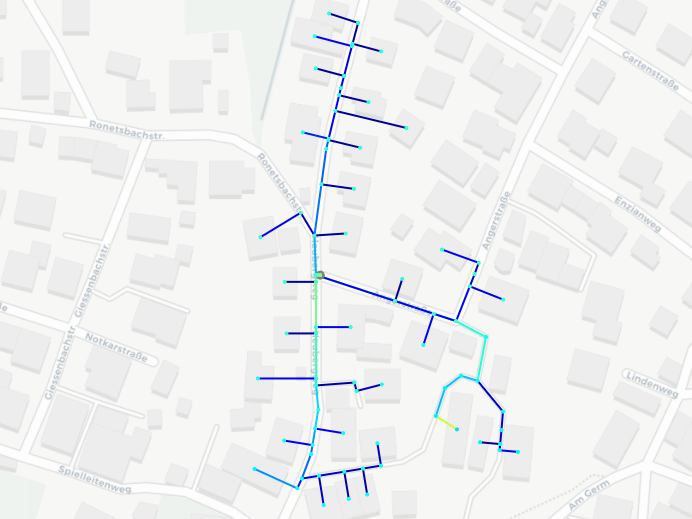} }}\qquad
    \subfloat[\centering Medium-voltage topology]{{\includegraphics[width=.4\linewidth]{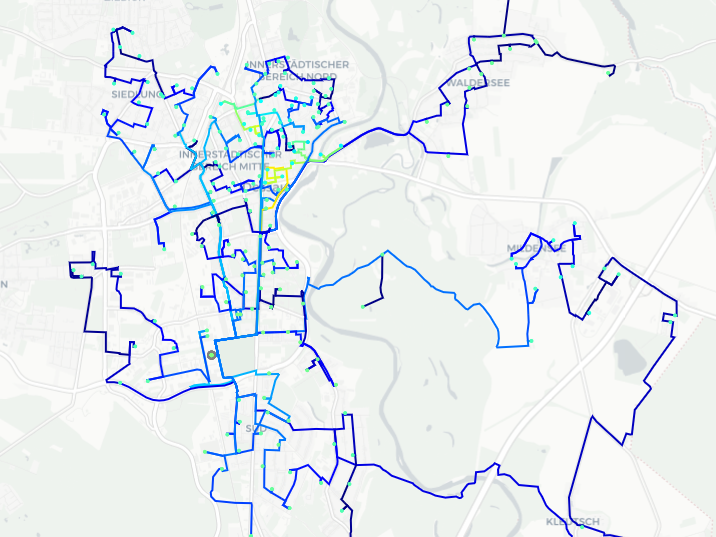} }}\caption{Exemplary topologies created by the grid generation process}\label{fig: topology_example_real}\end{figure}

For training, testing and validation of our approach we are modelling synthetic grids on the basis of open map data. Further, we model a geo-referenced supply task to determine low-voltage (LV) and medium-voltage (MV) grids representing realistic distribution grid models for the medium and low-voltage levels.\label{modelling distribution grids} Prior to the generation of realistic distribution grids, a supply task has to be assigned to a region by assigning commercial and residential loads as well as generation capacities to georeferenced structures. On such a regionalized supply task several methods exist to model the electrical distribution grid \cite{Amme.2018, Kays.2017, Sprey.2021}.\\
To create a realistic training dataset for our approach we chose a similar method to create multiple grid models for the medium-voltage level.
After the assignment of the supply task, we define objects which are then connected to the electrical grid. These are buildings, points of interest and the locations of electrical power equipment. The assignment process is based on information provided by the OpenStreetMap (OSM) database \cite{OpenStreetMap} resulting in a set of individual, georeferenced connection objects. All these objects include properties like electrical demand and generation, power rating and annual power consumption. 
The previously defined connection objects are then to be connected to the modelled power grid based on their electrical rating.\\
The topological structure of the electrical power grids generated is based on the road network which is also provided by OSM since power lines are historically placed and routed along roads. The modelling of the medium-voltage grids is based on the prior modelling of the corresponding low-voltage grids. In low-voltage, the connection objects are connected to the road network via the shortest paths. A graph partitioning algorithm subsequently partitions the supplied areas for each secondary substation. Originating at the secondary substation, the capacitated minimum spanning tree is formed along the road network to form radial grids, that connect all connection objects to the secondary substation. Based on the primary substations extracted from OSM and the locations of the secondary substations, the medium-voltage grids are modelled. With a capacitated vehicle routing heuristic the secondary substations as well as connection objects directly connected to medium-voltage are then connected to the primary substation forming open electrical ring grids.
The electrical ratings of all grid assets are determined by an assignment of typical asset types with an expansion heuristic to assure congestion-free operation of the grid based on a power flow analysis \cite{TRAGESER2022108217}.
An example of the created grid models is presented in Fig. \ref{fig: topology_example_real}.

\subsection{Data Set Generation}
For the generation of training, test and validation data sets, the data set generation process can be applied to each topology. This process can be used for single as well as multiple topologies. 
Each grid topology is used in the process to generate multiple case files on which the power flow calculation can be performed. To achieve the assignment of diverse but realistic supply tasks to each topology a consistent procedure is used.

\begin{figure}[h]
    \centering
    \subfloat[\centering Low-voltage topology]{{\includegraphics[width=.4\linewidth]{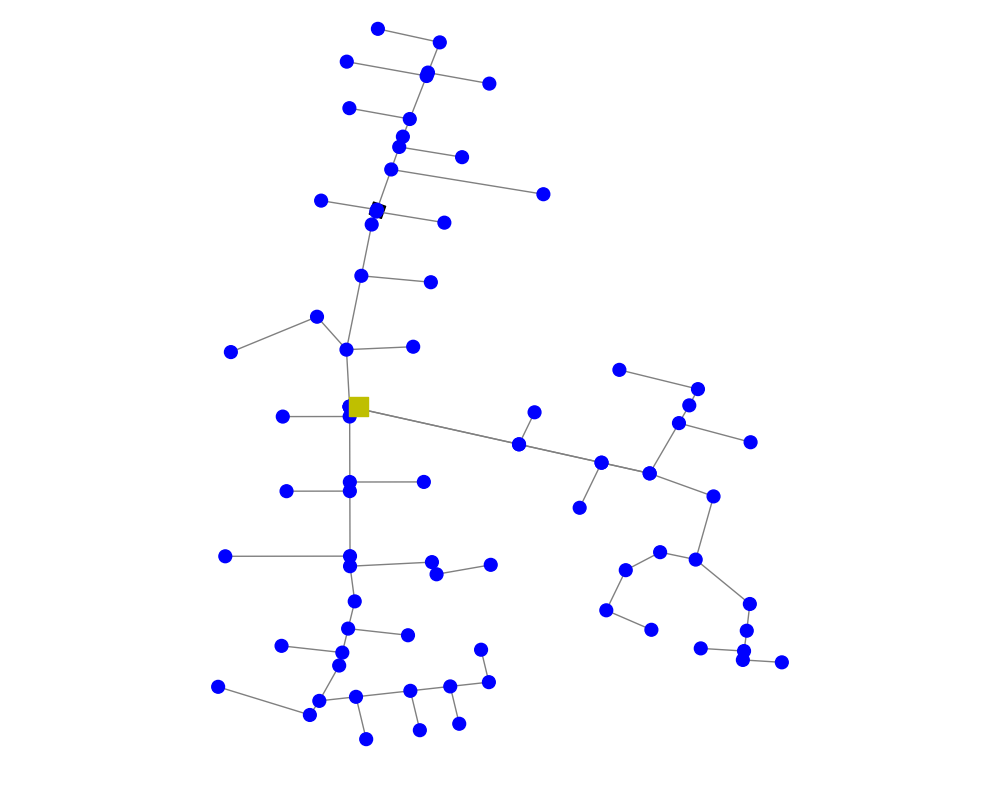} }}\qquad
    \subfloat[\centering Medium-voltage topology]{{\includegraphics[width=.4\linewidth]{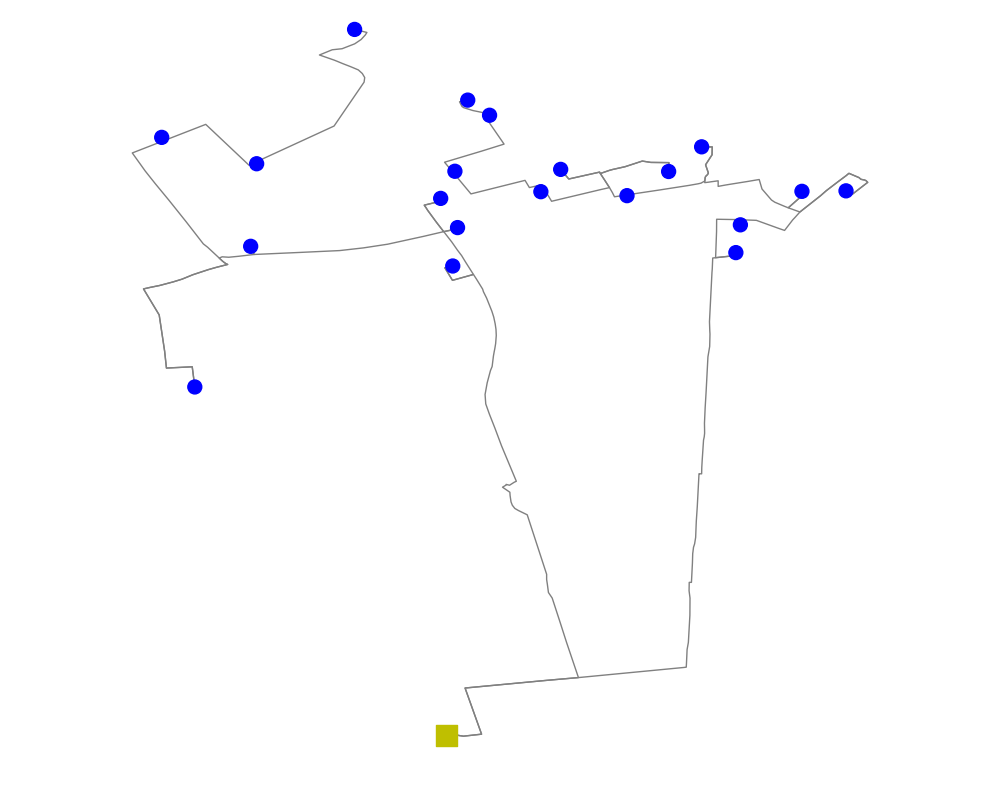} }}\caption{Exemplary topologies processed in the data set generation procedure}\label{fig: topology_example}\end{figure}

\label{topology check}
Initially, a topology check is performed to ensure that each topology is supplied by a single slack node. If a slack node is supplying more than one medium-voltage ring, the initial topology is split into separate topologies. As usable topologies for training, just single voltage level topologies with one slack node are used.
\label{node assignment}
In the next step, the node types as presented in Table \ref{tab:bus_types} (PQ and PV nodes) are randomly assigned to the busses on the topologies. The slack node was not altered and is located at the primary substation in each topology. 
\label{load and generation assignment}
After the assignment of the node types, the corresponding required values are set as well.
\label{power flow convergence}
To ensure power flow convergence all assigned values are restricted to a realistic range of values based on the analysis of load and generation time series provided by SimBench \cite{SteffenMeinecke.2020}.
As a result, a data set consisting of pairs of unsolved power flow cases and matching solutions for each topology and supply task is created. 
An example of the created grid graphs representing the grid models is presented in Fig. \ref{fig: topology_example}.

A resulting training data set is then provided to the GNN training procedure.
For interoperability,  we chose the PYPOWER casefile format (ppc) to store the power flow data sets
and made our model compatible with this format. PYPOWER and the ppc are python translations of the MatLab-based MATPOWER package and the MATPOWER casefile (mpc), which is also regularly used for power flow simulations \cite{Zimmerman.2011}.

\subsection{Loss Function} 
For our physics-aware training approach, we utilize a loss function that evaluates the solution computed by the Graph Neural Network.
Our loss function for each node \(\mathbf{v} \in \mathbf{V}\) consists of the power balance for real power and reactive power. Therefore, a power flow is correct when the physical constraints described by the AC power flow equations are met with equality, resulting in
\begin{equation*}
    L_{P_{\mathbf{v}}}  \coloneqq P_{\mathbf{v}} - \sum_{\mathbf{w} \in \mathbf{V}} |V_\mathbf{v}||V_\mathbf{w}|(G_{\mathbf{v}\mathbf{w}}\cos\vartheta_{\mathbf{v}\mathbf{w}}+B_{\mathbf{v}\mathbf{w}}\sin\vartheta_{\mathbf{v}\mathbf{w}})
\end{equation*}
\begin{equation*}
    L_{Q_{\mathbf{v}}}  \coloneqq Q_{\mathbf{v}} - \sum_{\mathbf{w} \in \mathbf{V}} |V_\mathbf{v}||V_{\mathbf{w}}|(G_{\mathbf{v}\mathbf{w}}\sin\vartheta_{\mathbf{v}\mathbf{w}}-B_{\mathbf{v}\mathbf{w}}\cos\vartheta_{\mathbf{v}\mathbf{w}})
\end{equation*}
where $V_{\mathbf{v}}$ and $\vartheta_{\mathbf{v}}$ denote the voltage magnitude and angle at node $\mathbf{v}$, while $P_{\mathbf{v}}$ and  $Q_{\mathbf{v}}$ denote active and reactive power respectively.
The goal of the GNN is to minimize \(L_{P_{\mathbf{v}}}\) and \(L_{Q_{\mathbf{v}}}\).
This is done during training by updating the internal weights of the neural network.
However, a GNN cannot use individual loss terms for every node, therefore the losses are summed to the following loss formulation:
\begin{equation*}
     L_{\text{total}} \coloneqq \sum_{\mathbf{v} \in \mathbf{V}} |L_{P_{\mathbf{v}}}| + | L_{Q_{\mathbf{v}}}|
\end{equation*}

Empirically, \(L_{\text{total}}\) has shown slow convergence.
Better results were obtained with the following variant of the loss function:
\begin{equation*}
L_{\text{train}} \coloneqq \ln\big(1 + \sum_{\mathbf{v} \in \mathbf{V}} L_{P_{\mathbf{v}}}^2 + L_{Q_{\mathbf{v}}}^2\big)
\end{equation*}
We report the loss for interpretability as loss per node \(\nicefrac{L_{\text{total}}}{|\mathbf{V}|}\) to obtain a metric that is independent of the size of any given topology. 
It is then verified that an accurate solution to the power flow problem results in a loss of exactly zero for $L_{total}$ as well as $L_{train}$.
We use this loss function in training with a variety of grid supply tasks.
For batching, we use the approach by \cite{DBLP:conf/iclr/KipfW17}, where all graphs in a batch are merged in a new graph such that there exists not any edge between two old graphs in the new graph. So we take the disjoint union of the graphs.
Further, all edges of the old graphs are preserved in the new graph.
By this, we obtain one unified graph that has as many connected components as we have graphs batched together.
This technique applies the loss function on any graph individually and is then summed up over the whole batch.

To train our neural network architecture, we chose the ADAM optimizer \cite{adam} and selected hyperparameters like batch size, layer sizes and learning rate with a conducted hyperparameter sweep.

\subsection{Architecture}
Our architecture is based on a randomized, recurrent GNN inspired by \cite{DBLP:journals/frai/TonshoffRWG20}. The architecture performs message passing on a graph $\mathbf{G}'$ that is constructed by representing each node or edge of the electrical graph $\mathbf{G}$ with a node. Further, undirected edges in $\mathbf{G}$ are split into two opposing directed edges in $\mathbf{G}'$. As a result, every edge in the resulting graph connects an electrical node to an electrical edge or vice versa. This enables the network to maintain hidden states for nodes and edges.  

\begin{figure}[h]
    \centering
    \includegraphics[width=0.4\textwidth]{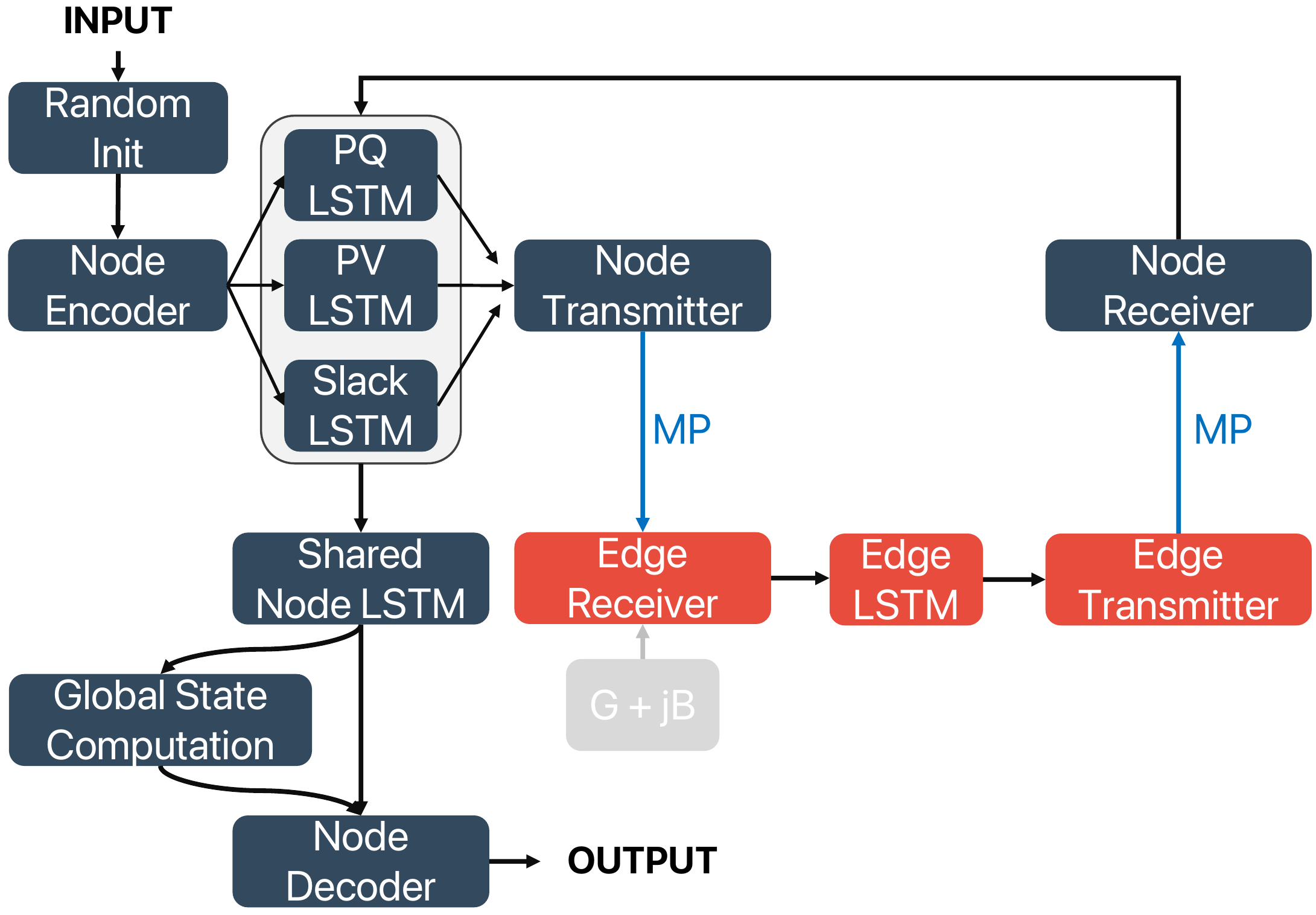}
    \caption{Simplified recurrent, randomized GNN architecture operating on the modified graph.}
    \label{fig:Architecture}
\end{figure}

The randomization takes place by adding noise sampled from a Gaussian mixture model to the decision variables (see unknown variables in Table~\ref{tab:bus_types}) before feeding the data to the message passing GNN. This is done to learn how to solve the power flow problem from many different starting points. 
The architecture consists of three main building blocks: An encoder, the recurrent message passing stack and a decoder. The idea is to transform the node variables into a high dimensional latent space via the encoder before iteratively improving the approximation of a power flow solution via message passing. Ultimately, we decode the node representations back to the node variables at every process step. Because of the recurrent structure, we obtain one solution candidate for each iteration of message passing, allowing us to select the best candidate by comparing their losses. The architecture is visualized in Figure \ref{fig:Architecture}.

The encoder and decoder are multi-layer perceptrons (MLP) using leaky ReLU activations \cite{maas2013rectifier}.
The encoder receives one feature vector per node, producing an abstract representation of node features.
The input of the GNN consists of the node variables $P$, $Q$, $V$, $\vartheta$ plus additional engineered features based on these variables.
Together with a one-hot encoding signifying the node type (PQ, PV or Slack), these node features are fed to the encoder.
The recurrent stack then processes the encoder's embedding of length $d=150$.
Long Short-Term Memory (LSTM) \cite{DBLP:journals/neco/HochreiterS97} cells update the embeddings for electrical nodes and edges every iteration.
The LSTMs get as input the hidden state of the last layer of the encoder, the results of the message-passing and their own cell state.
A different cell is used for each node type. 
The message-passing procedure consists of multiple steps: For each edge, messages are computed via an MLP from the connected nodes using their node states. Next, these embeddings are concatenated with the corresponding edge's impedance information ($G$ and $B$) before being fed to the edge receiver module, another MLP. The result is used to update the edge state using an LSTM cell. Finally, we perform the same procedure backwards from the updated edge states: Each edge computes a message via the transmitter MLP before these messages are passed to the nodes, whereas another MLP receives them before they are used to update the node states. After all, message passing iterations, all node states are passed through one more LSTM cell. The decoder receives these hidden states of all electrical nodes together with a global embedding (the mean of all node embeddings) project back into the  space of $P, Q, V,$ and $\vartheta$.

 \begin{table}[b]
    \centering
        \caption{Mean Statistics of the Data Sets' NR-Solutions}\begin{threeparttable}
    \begin{tabular}{lccc}
\toprule          & Synthetic MV & SimBench & Real MV \\
\midrule
        Network Size          & 17.41    & 100    & 120  \tnote{*}        \\
        PQ Node Count         & 16.33 &     97 & 118 \\  
        PV Node Count         & 0.081  &2 & 0  \\
        $P_{PQ}$ [MW]         & 0.148    & -0.0638  & 0.153    \\
        $Q_{PQ}$ [MVAr]       & 0.009    & 0.010    & 0.010      \\
        $P_{PV}$ [MW]         & -0.507   & -1.587   & -        \\
        $Q_{PV}$ [MVAr]       & -2.206   & 13.627   & -          \\
        $P_{Slack}$ [MW]      & -2.430   & 7.059    & -9.648   \\
        $Q_{Slack}$ [MVAr]    & 0.300    & -27.329  & 0.059      \\
        $V$ [pu]              & 0.99305  & 1.020    & 0.940    \\
        $\vartheta$ [deg]     & -0.332   & 1.26     & -1.664    \\
\bottomrule
    \end{tabular}
    \begin{tablenotes}
    \item[*] separable into two disjunct grids with 100 and 20 nodes, respectively
    \end{tablenotes}
    \end{threeparttable}

    \label{tab:dataset_stats}
\end{table}
\section{Experiments}

For our experiments, we created multiple data sets for the training process. To get a realistic set of grid topologies and respective supply tasks, the node types in each topology were preset to a realistic distribution of PV and PQ nodes with single slack nodes. 
The statistics of the different data sets used in our experiments are displayed in Table \ref{tab:dataset_stats}. Presented are the mean values at the nodes of all power flow problem instances in our data sets after computing the power flow solutions using the \NR{} (NR) algorithm.

We apply the trained GNN to unseen data from different sources to evaluate our proposed method.
Firstly, we evaluate our proposed method on new grid topologies with random supply tasks generated by the same process as the training data using open data and SimBench time series data (synthetic MV).
Secondly, we apply the GNN to a publicly available medium voltage grid topology taken from the SimBench data set \cite{SteffenMeinecke.2020}, with randomly applied supply tasks.
Lastly, we perform fine-tuning (continued training on a more specific data set) on the \emph{General GNN} model to specialise the model to the domain of a SimBench grid topology as well as a real grid model (Real MV) provided by Schleswig-Holstein-Netz AG before evaluating it on new supply tasks on the identical topologies.
We use a real-world medium voltage grid to test our method under realistic conditions. 
The benefits of such fine-tuned models are discussed in Section \ref{sec:Discussion}.
The GNN's accuracy in terms of the power flow loss is compared to the Graph Neural Solver (GNS) \cite{Donon.2020} and the \NR{} algorithm.
At last, we compare the inference computation time of our method and the run time of a \NR{} implementation.

\begin{table}[b]
    \centering
    \caption{Performance as Mean (and Median) Loss per Node in MVA for different MV grid models} \begin{tabular}{lccc}
            \toprule & Synthetic MV & SimBench & Real MV \\  
            \midrule General GNN & 0.152 (0.118) & 5.093 (5.221) & 0.323 (0.320)  \\
            FT SimBench & - & 0.112 (0.109) & - \\
            FT Real MV & - & - & 0.169 (0.177)\\
            \midrule
            GNS & 0.7318 & 0.9140 & 1.8276  \\  
            \NR{} & 0.0003 & 0.0003  & 0.0095  \\
\bottomrule
    \end{tabular}
    \label{tab:results}
\end{table}

Table \ref{tab:results} contains the average loss per node in MVA when evaluating our model using withheld data from the different data sets. 
The general GNN model was trained using synthetically generated medium voltage grids based on open data. This data set consisted of 42,500 power flow problem instances in total. 
The corresponding test set (Synthetic MV) consists of over 10,000 instances which are based on different topologies compared to the training set.
The fine-tuned models (FT) are trained to specialise in a specific grid topology and are then evaluated on unseen supply tasks on the same topology. 
While the SimBench data set contains 10,000 supply tasks, the Real MV data set consists of 20,000. 
1,000 problem instances were separated for testing for each of these specialised data sets.
The supply tasks were randomly generated using the same process as with the synthetic MV data.
To evaluate the model's ability to generalise over unseen topologies and supply tasks.

\begin{figure}
    \centering    \includegraphics[width=\columnwidth]{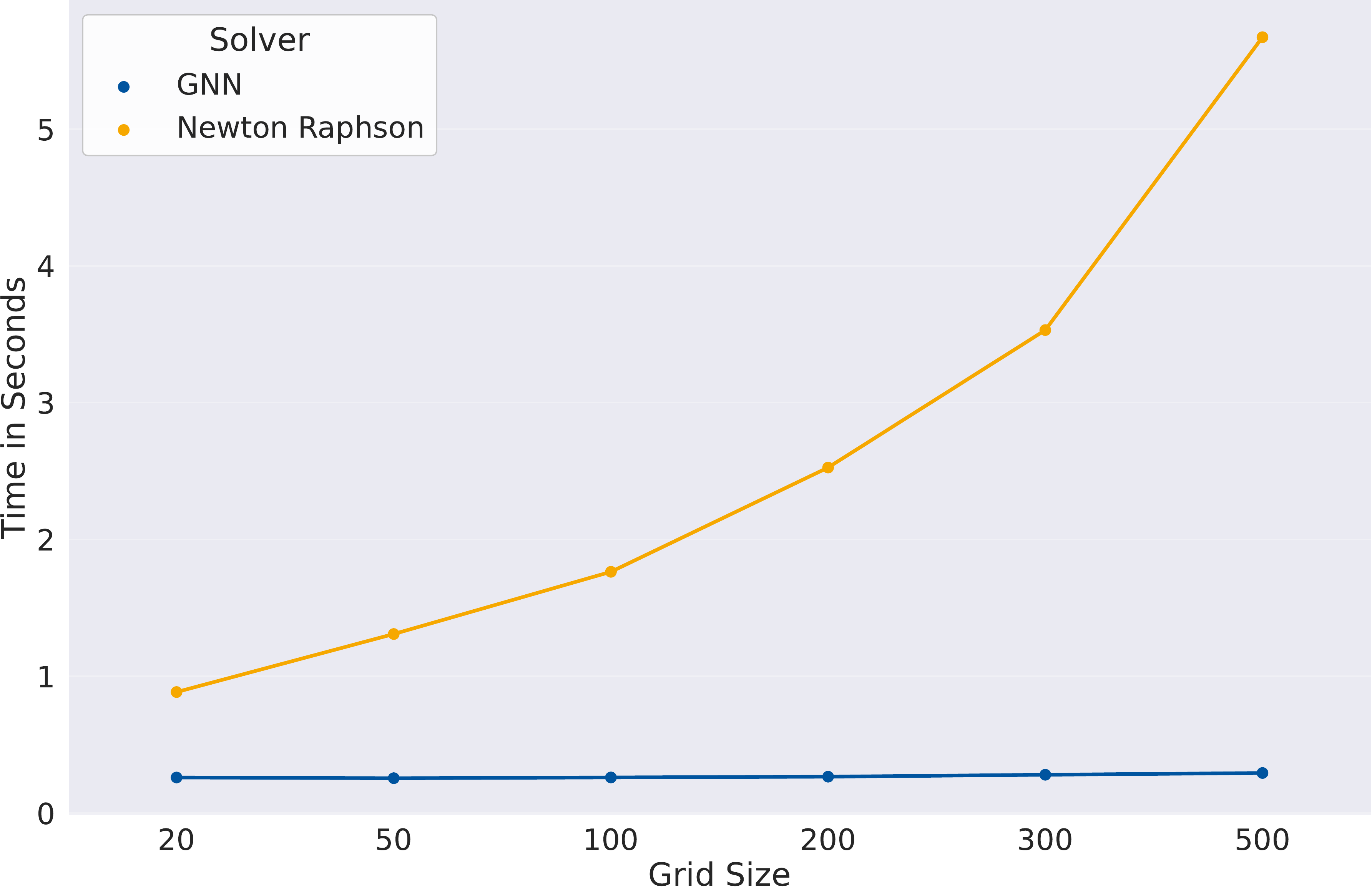}
    \caption{Evaluation times of Model}
    \label{fig:computation_times}
\end{figure}

The loss used here can be interpreted as the average amount of power (real and reactive) that is lost at each node by the algorithm, meaning a perfect solution yields a value of precisely zero.
Due to our recurrent architecture, we get We test each model on all the available data sets tony potential solutions for a single forward pass through our neural network. 
Additionally, we use random initialization for the decision variables at each node. 
Consequently, one can improve the method's accuracy by repeatedly evaluating the network on the same problem instance using different initializations and picking the best attempt afterwards.  
The results in Table \ref{tab:results} are based on applying our method with 50 recurrent iterations and 10 random initializations.
Our method selects the best of these 500 attempts by choosing the solution with the lowest loss. 
Note that this can be done without the knowledge of a perfect solution.
The results of GNS were achieved by training directly on the datasets Synthetic MV, SimBench and Real MV, 

At last, we compared the evaluation times of our model with the computation time of the \NR{} method.
In this regard, we used a set of grids with different sizes.
The data set was split into chunks of each 50 grids with sizes up to 20, 50, 100, 200, 300 and 500 busses.
In Figure \ref{fig:computation_times}, we depict the mean computation times of each bucket for our model as a single batch and \NR{}.
It can be observed that the GNN has a constant computation time for all grid sizes in the experiment, while \NR{} highly depends on the grid size.

\subsubsection{Implementation}
Our GNNs were implemented using PyTorch \cite{PyTorch} and PyTorch-Geometric \cite{TorchGeo}. 
Each model was trained on an Nvidia Tesla-V100 GPU with 16 GB of memory for 5000 epochs, while the FT-models were initialized with weights from the \emph{General GNN} training. 
For the \NR{} algorithm, we used PYPOWER's\footnote[1]{https://github.com/rwl/PYPOWER/} implementation.
For the GNS, we used the implementation provided through GitHub\footnote[2]{https://github.com/bdonon/GraphNeuralSolver}.

 \section{Discussion}
\label{sec:Discussion}
Our proposed model learns to compute AC Power Flow on medium-voltage grids in real-world scenarios.
This means our methods can be applied to any grid topology, including meshed topologies, as they are found in high-voltage and extra-high-voltage grids.
In general, our approach is independent of the voltage level.
In particular, our GNN architecture can represent the different bus types required to model the power flow problem as it occurs in medium and low-voltage grids.
Unlike the first approach\cite{Donon.2019}, we can handle line impedances which makes our method applicable to any real-world grid.
However, in \cite{Donon.2020}, the authors modified their model to handle line impedance, too.
Once our GNN has been trained on various grid topologies, the model can solve the power flow problem on any electrical grid, even if its underlying topology has not been used in training. 

The main goal of our approach was to achieve a good generalization performance.
We wanted to perform well over various supply tasks on the same topology but also aimed at generalizing over a large spectrum of topologies.
We demonstrate the generalization capabilities by evaluating the \emph{General GNN} model on the different data sets denoted in Table \ref{tab:results}.
The Synthetic MV data test set consists of different topologies compared to its training data set.
Therefore, the generalization over different topologies is shown by the evaluation of \emph{General GNN} (i) on Synthetic MV and (ii) further on SimBench and Real MV, which borrow their topologies from external data sources.
Moreover, we even generalize to significantly larger topologies as demonstrated by (ii), as can be seen in combination with the mean grid sizes found in Table \ref{tab:dataset_stats}.
In these cases, we observe a drop in loss per node, which we attribute to the size of those topologies, as those require information to travel further in the grid topology.
When evaluating the General GNN model on the SimBench data, one observes a significant loss compared to the real-world grid and the Synthetic MV data.
We attribute this to the unique properties of the SimBench grid, which has many static generators, yielding many PQ nodes that inject power into the grid.
According to our training data, this is an improbable scenario, as can also be seen by comparing the mean power values of the data sets in Table \ref{tab:dataset_stats}.
The fact that the performance on the even larger real-world medium voltage grid is much more in line with our loss on the Synthetic MV grid supports this argument.

The experiments with fine-tuned models demonstrate that performance improves when training specifically for a given topology. 
In this case, the model only needs to learn to solve the power flow problem for multiple supply tasks without generalising beyond the training data's topologies.
Applying the model in this non-generalized way is helpful for continuous (real-time) calculation of the status of a particular grid that is well-monitored in terms of power generation and demand (e.g. smart metering) or voltage levels (low-cost monitoring). As a second use case, time series simulations for a fixed topology can be sped up significantly. Especially for applying a fine-tuned GNN model in grid planning or operation approaches, the run-time can be considerably enhanced compared to a Newton-Raphson power flow.
Regarding the solution accuracy presented in Table \ref{tab:results}, our method yields promising results.
When one initializes the Synthetic MV data set with a so-called flat start, setting all unknown voltages to $1.0$ p.u. and all other unknown values (see Table \ref{tab:bus_types}) to zero, one gets a mean loss per node of $0.96$ for this data set.  
Therefore, our method does improve the power flow balance significantly compared to this baseline. 
However, while our method does not yet achieve the Newton-Raphson solution in terms of accuracy, it outperforms the Graph Neural Solver without fine-tuning.
Notably, the performance of the GNS reduces with the size of the grids.

A crucial part of our approach is creating the training data using synthetic grid data. 
The GNN training showed to be very dependent on the distribution of node types chosen in this process. 
We aimed to create a realistic training data set and chose the node type distribution accordingly. 
When the distribution of node types at test time is different compared to the training set, an increase in loss per node is observed at evaluation, especially when there are many PV nodes.    \section{Conclusion}
Our GNN architecture is a suited method to approximate AC Power Flow solutions under realistic constraints in a fast and scalable way by exploiting the parallel computations of modern GPUs.
In contrast to previous approaches, we train and evaluate a variety of realistic distribution grids. With  the training data generation approach, realistic power flow results have been determined on a variety of grid topologies, which deviates from existing methods and can thus act as a basis for comparing different ML models to determine the power flow.

Once trained, our presented model can generalize over different grid topologies and supply tasks.
By fine-tuning a fixed topology and multiple random initializations, we can improve the accuracy of the power flow solution provided by the GNN.
Our experiments showed that we achieved higher accuracy than the approach presented in \cite{Donon.2020}.
In real-time applications where computation time is critical, our method can enhance the provision of power flow solutions.
The \NR{} algorithm should be considered when high accuracy is required. Still, it might be improved by warm-starting the algorithm with the solution provided through the GNN architecture and thus reducing convergence steps.

In future work, we plan to integrate our GNN with reinforcement learning.
With this, we expect to reach a higher accuracy as we can adjust the node states only partially while designing a custom reward function.

\begin{thebibliography}{10}
\providecommand{\url}[1]{#1}
\csname url@samestyle\endcsname
\providecommand{\newblock}{\relax}
\providecommand{\bibinfo}[2]{#2}
\providecommand{\BIBentrySTDinterwordspacing}{\spaceskip=0pt\relax}
\providecommand{\BIBentryALTinterwordstretchfactor}{4}
\providecommand{\BIBentryALTinterwordspacing}{\spaceskip=\fontdimen2\font plus
\BIBentryALTinterwordstretchfactor\fontdimen3\font minus
  \fontdimen4\font\relax}
\providecommand{\BIBforeignlanguage}[2]{{\expandafter\ifx\csname l@#1\endcsname\relax
\typeout{** WARNING: IEEEtran.bst: No hyphenation pattern has been}\typeout{** loaded for the language `#1'. Using the pattern for}\typeout{** the default language instead.}\else
\language=\csname l@#1\endcsname
\fi
#2}}
\providecommand{\BIBdecl}{\relax}
\BIBdecl

\bibitem{Donon.2019}
B.~Donon, B.~Donnot, I.~Guyon, and A.~Marot, ``Graph neural solver for power
  systems,'' in \emph{2019 International Joint Conference on Neural Networks
  (IJCNN)}.\hskip 1em plus 0.5em minus 0.4em\relax IEEE, 2019, pp. 1--8.

\bibitem{Donon.2020}
B.~Donon, R.~Cl{\'e}ment, B.~Donnot, A.~Marot, I.~Guyon, and M.~Schoenauer,
  ``Neural networks for power flow: Graph neural solver,'' \emph{Electric Power
  Systems Research}, 2020.

\bibitem{Liao.2022}
W.~Liao, B.~Bak-Jensen, J.~{Radhakrishna Pillai}, Y.~Wang, and Y.~Wang, ``A
  review of graph neural networks and their applications in power systems,''
  \emph{Journal of Modern Power Systems and Clean Energy}, vol.~10, no.~2, pp.
  345--360, 2022.

\bibitem{Owerko.2019}
\BIBentryALTinterwordspacing
D.~Owerko, F.~Gama, and A.~Ribeiro, ``Optimal power flow using graph neural
  networks.'' [Online]. Available: \url{http://arxiv.org/pdf/1910.09658v1}
\BIBentrySTDinterwordspacing

\bibitem{Jeddi.2021}
A.~B. Jeddi and A.~Shafieezadeh, ``A physics-informed graph attention-based
  approach for power flow analysis,'' in \emph{2021 20th IEEE International
  Conference on Machine Learning and Applications (ICMLA)}.\hskip 1em plus
  0.5em minus 0.4em\relax IEEE, 2021, pp. 1634--1640.

\bibitem{Hansen.2021}
\BIBentryALTinterwordspacing
J.~B. Hansen, S.~N. Anfinsen, and F.~M. Bianchi, ``Power flow balancing with
  decentralized graph neural networks,'' \emph{CoRR}, vol. abs/2111.02169,
  2021. [Online]. Available: \url{https://arxiv.org/abs/2111.02169}
\BIBentrySTDinterwordspacing

\bibitem{Hu.2021}
X.~Hu, H.~Hu, S.~Verma, and Z.-L. Zhang, ``Physics-guided deep neural networks
  for power flow analysis,'' \emph{IEEE Transactions on Power Systems},
  vol.~36, no.~3, pp. 2082--2092, 2021.

\bibitem{Chow.2020}
J.~H. Chow and J.~J. Sanchez-Gasca, \emph{Power system modeling, computation,
  and control}.\hskip 1em plus 0.5em minus 0.4em\relax Hoboken NJ: {Wiley-IEEE
  Press}, 2020.

\bibitem{DBLP:conf/iclr/KipfW17}
\BIBentryALTinterwordspacing
T.~N. Kipf and M.~Welling, ``Semi-supervised classification with graph
  convolutional networks,'' in \emph{5th International Conference on Learning
  Representations, {ICLR} 2017, Toulon, France, April 24-26, 2017, Conference
  Track Proceedings}.\hskip 1em plus 0.5em minus 0.4em\relax OpenReview.net,
  2017. [Online]. Available: \url{https://openreview.net/forum?id=SJU4ayYgl}
\BIBentrySTDinterwordspacing

\bibitem{DBLP:conf/aaai/0001RFHLRG19}
\BIBentryALTinterwordspacing
C.~Morris, M.~Ritzert, M.~Fey, W.~L. Hamilton, J.~E. Lenssen, G.~Rattan, and
  M.~Grohe, ``Weisfeiler and leman go neural: Higher-order graph neural
  networks,'' in \emph{The Thirty-Third {AAAI} Conference on Artificial
  Intelligence, {AAAI} 2019}, 2019, pp. 4602--4609. [Online]. Available:
  \url{https://doi.org/10.1609/aaai.v33i01.33014602}
\BIBentrySTDinterwordspacing

\bibitem{DBLP:conf/iclr/XuHLJ19}
\BIBentryALTinterwordspacing
K.~Xu, W.~Hu, J.~Leskovec, and S.~Jegelka, ``How powerful are graph neural
  networks?'' in \emph{7th International Conference on Learning
  Representations, {ICLR} 2019, New Orleans, LA, USA, May 6-9, 2019}.\hskip 1em
  plus 0.5em minus 0.4em\relax OpenReview.net, 2019. [Online]. Available:
  \url{https://openreview.net/forum?id=ryGs6iA5Km}
\BIBentrySTDinterwordspacing

\bibitem{DBLP:journals/tnn/WuPCLZY21}
\BIBentryALTinterwordspacing
Z.~Wu, S.~Pan, F.~Chen, G.~Long, C.~Zhang, and P.~S. Yu, ``A comprehensive
  survey on graph neural networks,'' \emph{{IEEE} Trans. Neural Networks Learn.
  Syst.}, vol.~32, no.~1, pp. 4--24, 2021. [Online]. Available:
  \url{https://doi.org/10.1109/TNNLS.2020.2978386}
\BIBentrySTDinterwordspacing

\bibitem{DBLP:conf/icml/GilmerSRVD17}
\BIBentryALTinterwordspacing
J.~Gilmer, S.~S. Schoenholz, P.~F. Riley, O.~Vinyals, and G.~E. Dahl, ``Neural
  message passing for quantum chemistry,'' in \emph{Proceedings of the 34th
  International Conference on Machine Learning, {ICML} 2017, Sydney, NSW,
  Australia, 6-11 August 2017}, ser. Proceedings of Machine Learning Research,
  D.~Precup and Y.~W. Teh, Eds., vol.~70.\hskip 1em plus 0.5em minus
  0.4em\relax {PMLR}, 2017, pp. 1263--1272. [Online]. Available:
  \url{http://proceedings.mlr.press/v70/gilmer17a.html}
\BIBentrySTDinterwordspacing

\bibitem{DBLP:journals/frai/TonshoffRWG20}
\BIBentryALTinterwordspacing
J.~T{\"{o}}nshoff, M.~Ritzert, H.~Wolf, and M.~Grohe, ``Graph neural networks
  for maximum constraint satisfaction,'' \emph{Frontiers Artif. Intell.},
  vol.~3, p. 580607, 2020. [Online]. Available:
  \url{https://doi.org/10.3389/frai.2020.580607}
\BIBentrySTDinterwordspacing

\bibitem{Amme.2018}
J.~Amme, G.~Ple{\ss}mann, J.~B{\"u}hler, L.~H{\"u}lk, E.~K{\"o}tter, and
  P.~Schwaegerl, ``The ego grid model: An open-source and open-data based
  synthetic medium-voltage grid model for distribution power supply systems,''
  \emph{Journal of Physics: Conference Series}, vol. 977, p. 012007, 2018.

\bibitem{Kays.2017}
J.~Kays, A.~Seack, T.~Smirek, F.~Westkamp, and C.~Rehtanz, ``The generation of
  distribution grid models on the basis of public available data,'' \emph{IEEE
  Transactions on Power Systems}, vol.~32, no.~3, pp. 2346--2353, 2017.

\bibitem{Sprey.2021}
J.~M. Sprey, ``{Ermittlung des Netzausbaubedarfs anhand georeferenzierter
  Verteilnetzmodelle},'' Dissertation, {RWTH Aachen} and {Print Production M.
  Wolff GmbH}, 2021.

\bibitem{OpenStreetMap}
{OpenStreetMap contributors}, ``{Planet dump retrieved from
  https://planet.osm.org },'' \url{ https://www.openstreetmap.org }, 2017.

\bibitem{TRAGESER2022108217}
\BIBentryALTinterwordspacing
M.~Trageser, M.~Pape, K.~Frings, P.~Erlinghagen, M.~Kurth, C.~M. Vertgewall,
  A.~Monti, and T.~Busse, ``Automated routing of feeders in electrical
  distribution grids,'' \emph{Electric Power Systems Research}, vol. 211, p.
  108217, 2022. [Online]. Available:
  \url{https://www.sciencedirect.com/science/article/pii/S0378779622004266}
\BIBentrySTDinterwordspacing

\bibitem{SteffenMeinecke.2020}
{Steffen Meinecke}, {D{\v{z}}anan Sarajli{\'c}}, {Simon Ruben Drauz}, {Annika
  Klettke}, {Lars-Peter Lauven}, {Christian Rehtanz}, {Albert Moser}, and
  {Martin Braun}, ``Simbench---a benchmark dataset of electric power systems to
  compare innovative solutions based on power flow analysis,'' \emph{Energies},
  vol.~13, no.~12, p. 3290, 2020.

\bibitem{Zimmerman.2011}
R.~D. Zimmerman, C.~E. Murillo-Sanchez, and R.~J. Thomas, ``Matpower:
  Steady-state operations, planning, and analysis tools for power systems
  research and education,'' \emph{IEEE Transactions on Power Systems}, vol.~26,
  no.~1, pp. 12--19, 2011.

\bibitem{adam}
\BIBentryALTinterwordspacing
D.~P. Kingma and J.~Ba, ``Adam: {A} method for stochastic optimization,'' in
  \emph{3rd International Conference on Learning Representations, {ICLR} 2015,
  San Diego, CA, USA, May 7-9, 2015, Conference Track Proceedings}, Y.~Bengio
  and Y.~LeCun, Eds., 2015. [Online]. Available:
  \url{http://arxiv.org/abs/1412.6980}
\BIBentrySTDinterwordspacing

\bibitem{maas2013rectifier}
A.~L. Maas, A.~Y. Hannun, A.~Y. Ng \emph{et~al.}, ``Rectifier nonlinearities
  improve neural network acoustic models,'' in \emph{Proc. icml}, vol.~30,
  no.~1.\hskip 1em plus 0.5em minus 0.4em\relax Atlanta, Georgia, USA, 2013,
  p.~3.

\bibitem{DBLP:journals/neco/HochreiterS97}
\BIBentryALTinterwordspacing
S.~Hochreiter and J.~Schmidhuber, ``Long short-term memory,'' \emph{Neural
  Comput.}, vol.~9, no.~8, pp. 1735--1780, 1997. [Online]. Available:
  \url{https://doi.org/10.1162/neco.1997.9.8.1735}
\BIBentrySTDinterwordspacing

\bibitem{PyTorch}
\BIBentryALTinterwordspacing
A.~Paszke, S.~Gross, F.~Massa, A.~Lerer, J.~Bradbury, G.~Chanan, T.~Killeen,
  Z.~Lin, N.~Gimelshein, L.~Antiga, A.~Desmaison, A.~Kopf, E.~Yang, Z.~DeVito,
  M.~Raison, A.~Tejani, S.~Chilamkurthy, B.~Steiner, L.~Fang, J.~Bai, and
  S.~Chintala, ``Pytorch: An imperative style, high-performance deep learning
  library,'' in \emph{Advances in Neural Information Processing Systems 32},
  H.~Wallach, H.~Larochelle, A.~Beygelzimer, F.~d\textquotesingle
  Alch\'{e}-Buc, E.~Fox, and R.~Garnett, Eds.\hskip 1em plus 0.5em minus
  0.4em\relax Curran Associates, Inc., 2019, pp. 8024--8035. [Online].
  Available:
  \url{http://papers.neurips.cc/paper/9015-pytorch-an-imperative-style-high-performance-deep-learning-library.pdf}
\BIBentrySTDinterwordspacing

\bibitem{TorchGeo}
\BIBentryALTinterwordspacing
M.~Fey and J.~E. Lenssen, ``Fast graph representation learning with pytorch
  geometric,'' \emph{CoRR}, vol. abs/1903.02428, 2019. [Online]. Available:
  \url{http://arxiv.org/abs/1903.02428}
\BIBentrySTDinterwordspacing

\end{thebibliography}
\end{document}